\documentclass[10pt,twocolumn,letterpaper]{article}

\usepackage{cvpr}              %

\usepackage{graphicx}
\usepackage{amsmath}
\usepackage{amssymb}
\usepackage{booktabs}
\usepackage[utf8]{inputenc} %
\usepackage[T1]{fontenc}    %
\usepackage{multirow}
\usepackage{pifont}
\usepackage[accsupp]{axessibility}

\usepackage[pagebackref,breaklinks,colorlinks]{hyperref}

\usepackage[capitalize]{cleveref}
\crefname{section}{Sec.}{Secs.}
\Crefname{section}{Section}{Sections}
\Crefname{table}{Table}{Tables}
\crefname{table}{Tab.}{Tabs.}

\def\cvprPaperID{}
\def\confName{CVPR}
\def\confYear{2022}

\newcommand{\citep}[1]{\cite{#1}}
\newcommand{\citet}[1]{\cite{#1}}
\newcommand{\vect}[1]{\mathbf{#1}}

\newcommand{\cc}{\mathbf{c}}
\newcommand{\ff}{\mathbf{f}}
\newcommand{\ii}{\mathbf{i}}

\newcommand{\xx}{\mathbf{x}}
\newcommand{\xxd}{\mathbf{\dot x}}
\newcommand{\pp}{\mathbf{p}}

\newcommand{\vv}{\mathbf{v}}

\newcommand{\qq}{\mathbf{q}}
\newcommand{\qqd}{\mathbf{\dot q}}
\newcommand{\qqdd}{\mathbf{\ddot q}}
\newcommand{\kk}{\mathbf{k}}

\newcommand{\mat}[1]{\mathbf{#1}}
\newcommand{\CC}{\mathbf{C}}

\newcommand{\HH}{\mathbf{H}}
\newcommand{\RR}{\mathbb{R}}

\newcommand{\TT}{\mathbf{T}}

\newcommand{\MM}{\mathbf{M}}

\newcommand{\LL}{\mathbf{L}}

\newcommand{\bbeta}{\boldsymbol{\beta}}
\newcommand{\btheta}{\boldsymbol{\theta}}

\newcommand{\btau}{\boldsymbol{\tau}}

\newcommand{\myparagraph}[1]{\vspace{0.1em}\noindent\textbf{#1}}

\newcommand{\cmark}{\ding{51}}
\newcommand{\xmark}{\ding{55}}

\newcommand{\Eq}[1]{\eqref{#1}}
\newcommand{\Fig}[1]{fig.~\ref{#1}}
\newcommand{\FFig}[1]{Fig.~\ref{#1}}
\newcommand{\Tab}[1]{tab.~\ref{#1}}
\newcommand{\TTab}[1]{Tab.~\ref{#1}}
\newcommand{\Sec}[1]{\S\ref{#1}}

\begin{document}

\title{Differentiable Dynamics for Articulated 3d Human Motion Reconstruction}

\author{Erik Gärtner\textsuperscript{\rm 1,}\textsuperscript{\rm 2}\\
\and
Mykhaylo Andriluka\textsuperscript{\rm 1}\\
\and
Erwin Coumans\textsuperscript{\rm 1 }
\and
Cristian Sminchisescu\textsuperscript{\rm 1}
\and
\textsuperscript{\rm 1}\bf{Google Research}, \textsuperscript{\rm 2}\bf{Lund University}\\
{\tt\small erik.gartner@math.lth.se} \\ \vspace{-2mm}
{\tt\small \{mykhayloa,erwincoumans,sminchisescu\}@google.com}
}

\maketitle

\begin{abstract}
We introduce DiffPhy, a differentiable physics-based model for articulated 3d human motion reconstruction from video.
Applications of physics-based reasoning in human motion analysis have so far been limited, both by the complexity of constructing adequate physical models of articulated human motion, and by the formidable challenges of performing stable and efficient inference with physics in the loop. 
We jointly address such modeling and inference challenges by 
proposing an approach that combines a physically plausible body representation with anatomical joint limits, a differentiable physics simulator, and optimization techniques that ensure good performance and robustness to suboptimal local optima.
In contrast to several recent methods~\cite{RempeContactDynamics2020,PhysCapTOG2020,Xie_2021_ICCV}, our approach readily supports full-body contact including interactions with objects in the scene.
Most importantly, our model connects end-to-end with images, thus supporting direct gradient-based physics optimization by means of image-based loss functions. We validate the model by demonstrating that it can accurately reconstruct physically plausible 3d human motion from monocular video, both on public benchmarks with available 3d ground-truth, and on videos from the internet.

\end{abstract}

\section{Introduction}
\begin{figure*}[t]
  \centering
  \includegraphics[width=0.95\linewidth]{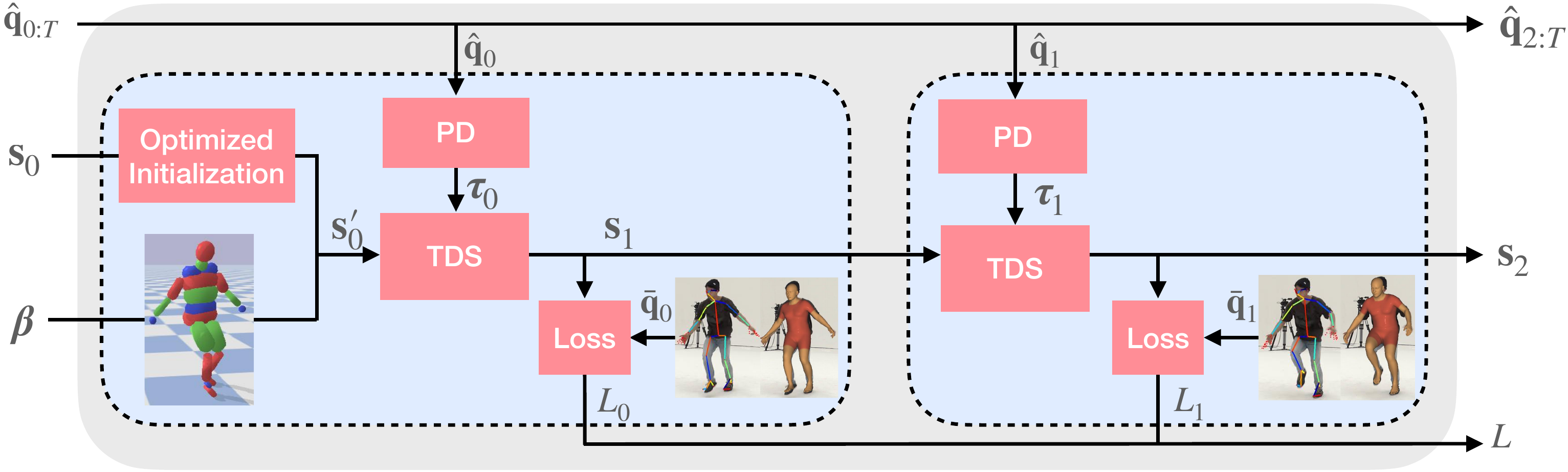}
  \vspace{-2mm}
  \caption{Overview of DiffPhy. Given kinematic estimates (described in \Sec{sec:kinematics}) of a subject's body shape $\bbeta$, the body's initial pose and velocity $\vect{s}_0$, and time-varying 3d poses $\bar\qq_{0:T}$ with detected 2d keypoints, our model reconstructs the motion in physical simulation, by minimizing a differentiable loss $L$ (see \Sec{sec:objectives}). DiffPhy optimizes the control trajectory $\hat\qq_{0:T}$ containing joint angle targets to PD controllers (\cf \Eq{eq:pd_controller}). In turn, the PD controllers compute a torque vector $\btau$, which actuates motors in the joints of the simulated body. DiffPhy integrates a full-featured differentiable simulator, TDS~\cite{heiden2021neuralsim} (described in \Sec{sec:differentiable_physics}), that supports complex contacts. Each subject is represented by means of a personalised physical model (see \Sec{sec:body_model}). In addition, we optimize the initial state (see \Sec{sec:optimized_initialization}), which makes DiffPhy robust to low quality initial estimates. The outputs are 3d pose estimates that align with visual evidence and respect physical constraints.}
  \label{fig:overview}
  \vspace{-4mm}
\end{figure*}

We seek to contribute to the development of physics-based methodology as one of the building blocks
in constructing accurate and robust 3d visual human sensing systems.
Incorporating the laws of physics into the visual reasoning process is appealing
as it promotes the plausibility of estimated motion and facilitates more efficient use of training
examples~\citep{belbute18neurips}.
We focus on articulated human motion as an epitome of a real-world prediction task that is both well
studied and challenging. Existing state-of-the-art approaches demonstrate
relatively high accuracy in terms of joint position estimation metrics~\cite{kanazawa2018end, Xiang_2019,kocabas20cvpr, Zanfir_2021_ICCV}. However, predictions can sometimes be physically implausible, even for simple motions such as walking and running. For instance, estimates can include unreasonably abrupt transitions in world space, or artifacts such as foot skating or non-equilibrium states~\citep{RempeContactDynamics2020,PhysCapTOG2020}.
Many methods are typically trained on large motion capture datasets and encounter difficulties when tested on motions not well represented in those training sets. 
Arguably, imposing some form of physics-based generally valid prior on the articulated motion estimates should
greatly improve the plausibility of results.

However, physics-based reasoning comes at the cost of substantial modeling and inference complexity.
Typically, physics-based  articulated estimation methods rely on rigid body dynamics
(RBD)~\cite{featherstone2007,stepien2013thesis}, a formulation that introduces many auxiliary
variables corresponding to forces acting at the body joints at each time step. Moreover, physical
contact results in non-smooth effects where small changes to model parameters might result in
substantially different motions. Therefore inferring physics
variables given the inherent uncertainty in monocular video, and under contact discontinuities, becomes significantly difficult, algorithmically
and computationally.
Despite such challenges, a number of recent methods successfully apply physics-based constraints for
articulated human motion estimation
\citep{alborno18cgf,PhysCapTOG2020,RempeContactDynamics2020,yuan2021simpoe}. One possibility to cope with modeling complexity, explored in recent work, is to simplify the
physics and model contacts only between the body and the feet
\cite{RempeContactDynamics2020,PhysCapTOG2020,Xie_2021_ICCV}. Others use auxiliary external forces
applied at the body to compensate for modeling error \citep{PhysCapTOG2020,yuan2021simpoe}.

In this paper, we aim to broaden the methodology for physics-based articulated
human motion estimation. Specifically, we demonstrate that we can successfully leverage recent progress in differentiable simulation \cite{heiden2021neuralsim,hu2019difftaichi,Werling2021FastAF} in order to incorporate physics-based constraints into the articulated 3d human motion reconstruction. Our approach, \emph{DiffPhy}, relies on gradient-based optimization, connects end-to-end with images, and does not require simplifying assumptions on
contacts or the introduction of external non-physical residual forces.

\section{Related work}
\myparagraph{Kinematics-based 3d Human Pose Estimation.}
The problem of monocular 3d pose estimation is usually addressed through end-to-end~\citep{mehta2017vnect, mehta2018single, zanfir2018monocular}, or two-stage~\citep{hossain2018exploiting, dabral2018learning} models where neural networks are used to predict 3d joint positions. This is an ill-posed problem due to depth ambiguities and occlusion. The networks are usually trained on vast pose datasets~\citep{h36mpami,AMASS:ICCV:2019, joo_total_motion_cap,vonMarcard2018} which usually supports good performance on poses previously observed during training. Several methods~\citep{zanfir2020neural,kocabas20cvpr, Zanfir_2021_ICCV} directly regress the parameters of statistical body models~\citep{SMPL:2015,xu2020ghum} (rather than 3d joint positions), including the subject's body shape as well as kinematic pose. 
The methods mentioned above take a purely visual inference approach to the problem and do not consider physics-based constraints. As observed by \citep{RempeContactDynamics2020}, this may cause  artifacts such as jitter, ground-penetration, foot sliding, or unnatural leaning~\cite{PhysCapTOG2020}.

\myparagraph{Physics-based 3d Human Pose Estimation.}
Recent work~\citep{RempeContactDynamics2020,PhysCapTOG2020,yuan2021simpoe,PhysAwareTOG2021,Shimada2021NeuralM3,Xie_2021_ICCV,luo2021dynamics,gartner2022trajectory} aims to increase realism, by using physics to regularize reconstruction. This aims to enforce physical constraints such as proper contact and dynamic coherence. In \citep{RempeContactDynamics2020} motion is reconstructed through optimization, but the method only accounts for collisions between the feet and the ground. Such simplifications are recurring in current approaches and limit the types of motions that can be reconstructed. In contrast, in this work, we use a full-featured physical simulator which supports contacts between all objects in the scene. PhysCap~\cite{PhysCapTOG2020} is a real-time optimization-based approach, where feet contact is pre-detected based on a neural network. During the physics-based inference, contacts are considered fixed and thherefore cannot be corrected or improved. Moreover, following \cite{yuan2020residual} the method uses non-physical ``residual forces'' which improve 3d joint reconstruction metrics at the cost of altered physical plausibility. Since we aim to increase the physicality of reconstructed motions, we avoid using any residual forces. \citep{yuan2021simpoe} follows on \citep{2018-TOG-deepMimic, ScaDiver} to learn a neural network that estimates torques to drive a model in the full-featured physical simulator MuJoCo~\citep{mujoco}.  However, MuJoCo is non-differentiable, hence the need to resort to expensive training using numerical gradients in a reinforcement learning setting. The method is trained for millions of steps using 3d ground-truth labels from a motion capture dataset, but the method's ability to generalize to in-the-wild is not demonstrated. Similarly to \citep{RempeContactDynamics2020},  the method assumes a known ground plane, whereas \emph{DiffPhy} estimates it. \citep{Shimada2021NeuralM3} integrates a simplified physics approach, dubbed ``physionical'', into a neural network that estimates joint torques and ground-reaction forces. Similarly to \citep{PhysCapTOG2020} they detect foot contact using a neural network predictor rather than by means of physical simulation. Most recently, \citep{Xie_2021_ICCV} introduced a method relying on a simplified physical formulation that makes it possible to refine 3D pose estimates well enough to train motion synthesis models based on that output. However, the method assumes a known ground plane, models only foot contact, and implements a simplified physical body scaled solely based on the estimated bone length rather than shape estimates. Finally, in our concurrent work \cite{gartner2022trajectory}, we perform physics-based human pose reconstruction of complex motions through trajectory optimization based on CMA-ES~\cite{Hansen2006} in the non-differentiable simulator Bullet~\cite{coumans2019pybullet}. This general approach uses a mature and full-featured simulator which, while capable, is slow due to costly black-box optimization. The method does not optimize the initial state of the body (see \Sec{sec:optimized_initialization}) together with the joint control variables, being more vulnerable to unfavorable initialisation. In summary, this work takes the novel approach of tightly integrating physics into the reconstruction process through a full-featured \emph{differentiable} physics model. As a result, DiffPhy supports complex full-body contacts, connects pixels-to-physics using end-to-end differentiable losses, supports personalised body models, does not resort to residual forces, and is robust to poor initialization. See \Tab{tab:competitors} for an overview of physics-based methods.

It is worth mentioning that, aside from physical simulation, there exist many other approaches to grounding the human pose estimates using e.g., inertial estimates from IMUs~\cite{yi2021imu}, scene constraints~\cite{zhang2020phosa,cao2020long}, and motion priors~\cite{rempe2021humor}.

\myparagraph{Differentiable Physics for Human Modeling.}
Physical simulation is a mature area with several
established simulation engines available \citep{coumans2019pybullet,mujoco,dart}. These engines
implement forward simulation but do not facilitate the computation of derivatives necessary for
efficient gradient-based optimization. These simulators are well-suited for training with gradient-free methods
such as reinforcement-learning or evolutionary algorithms and have been used for gradient-free
optimization of human motion models \citep{yuan2021simpoe,alborno18cgf,peng19dm}. More recently differentiable physics simulators have emerged \citep{carpentier2019pinocchio, Werling2021FastAF, heiden2021neuralsim, qiao2021efficient, brax2021github}. Applying these to human motion reconstruction is difficult due to noisy gradients~\cite{hu2019difftaichi,metz2021gradients}, and a non-convex objective function. We present a methodology using gradient-based local search with stochastic global optimization enabling the first use of a full-featured differentiable physics model~\cite{heiden2021neuralsim} for human pose reconstruction from video. Furthermore, we show that our approach is magnitudes faster than a purely sampling-based approach.

\begin{table}[bt]
\begin{center}
\scalebox{0.79}{
\begin{tabular}{l|c|c|c|c|c|c}
\textbf{Method} & \textbf{Body} & \textbf{Cont.} & \textbf{DP} & \textbf{Trained} & \textbf{$\TT_g$} & \textbf{No RF} \\
\hline
Rempe \etal \cite{RempeContactDynamics2020} & Fixed & Feet  & \xmark & Contacts & \xmark & \cmark \\
PhysCap~\cite{PhysCapTOG2020} & Fixed & Feet & \cmark & Contacts & \cmark & \xmark \\
SimPoE~\cite{yuan2021simpoe} & Adapt & Full & \xmark & Yes & \xmark & \xmark \\
Shimada \etal~\cite{Shimada2021NeuralM3} & Fixed & Feet & \cmark & Yes & \cmark & \xmark \\
Xie \etal \cite{Xie_2021_ICCV} & Fixed & Feet & \cmark & No & \xmark & \cmark \\
Dynamics \cite{gartner2022trajectory} & Adapt & Full & \xmark & Prior & \cmark & \cmark \\
DiffPhy & Adapt & Full & \cmark & No & \cmark & \cmark \\
\end{tabular}
}
\end{center}
\vspace{-4mm}
\caption{Feature comparison against other physics-based methods. \emph{Body} compares the type of physical body representation where ``adapt'' means individually constructed based on shape estimate, \emph{Cont.} column compares what type of contacts are supported, \emph{DP} whether the method uses a differentiable physical formulation, \emph{Training} if the physical inference requires training, \emph{$\TT_g$} compares if the ground plane is estimated (as opposed to assumed known), and \emph{No RF} if the method avoid non-physical residual forces. Only our method does not require any additional training and uses a full-featured differentiable physics formulation.
}
\label{tab:competitors}
\vspace{-2mm}
\end{table}
\begin{figure}[tb]
  \centering
  \includegraphics[width=1.0\linewidth]{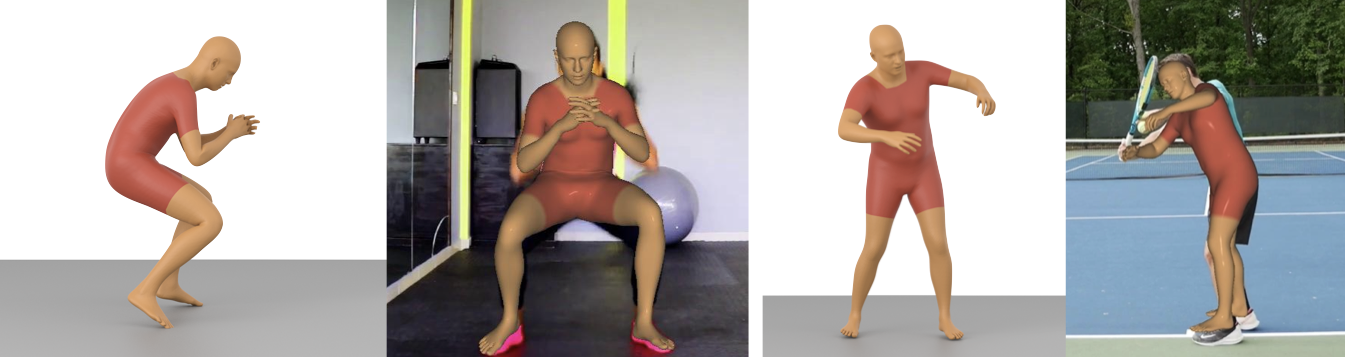}
  \vspace{-4mm}
  \caption{Qualitative results on two in-the-wild sequences. Sports and dynamic activities are rarely found in motion capture datasets.}
  \label{fig:pim_qunatitative}
  \vspace{-4mm}
\end{figure}

\section{Methodology}\label{sec:methodology}
This section presents our approach to reconstructing 3d human shape and motion from video with differentiable physics in the loop. Given a monocular video of a human subject, we use a kinematic neural network to estimate 2d body keypoints, the body shape, and 3d body poses. Since estimating 3d pose from monocular video is ill-posed, due to e.g. depth-ambiguities and occlusion~\cite{sminchisescu_cvpr03}, the kinematic 3d reconstructions may suffer from self-penetration, inconsistent translation, jitters, floating above the ground, and non-physical leaning~\citep{PhysCapTOG2020,RempeContactDynamics2020}. We, therefore, reconstruct the motion in physical simulation, by jointly accounting for both visual evidence and the constraints of physical simulation (e.g. collisions, gravity, and Newton's laws of motion). See \Fig{fig:overview} for an overview of our approach.

\subsection{Kinematic Initialization}\label{sec:kinematics}
Given a sequence of monocular images $\{I_i\}$, we assume a pinhole camera with intrinsics $\ii = [f_x, f_y, c_x, c_y]$ and constant camera extrinsics. We obtain the visual evidence used in our optimization objectives following the procedure introduced in \cite{gartner2022trajectory}. This relies on HUND~\citep{zanfir2020neural}, a 3d pose estimator that produces per-frame 2d keypoints $\bar \xx_i$ with confidence scores $\cc_i$, 3d body poses $\btheta_i$, and 3d body shape $\bbeta_i$, where $\btheta$ and $\bbeta$ are the GHUM~\citep{xu2020ghum} posing and shape parameters, respectively. 

Since HUND is a per-frame estimator, a temporally consistent shape is recovered by selecting the $N=5$ highest-scoring frames according to keypoint confidences. For these frames, HUND image losses~\citep{zanfir2020neural} are minimized using BFGS under the additional constraint of a constant shape, $\bbeta$, across all frames. In addition, \citep{gartner2022trajectory} introduces a final round of optimization where poses are updated under the time-consistent body shape and a temporal smoothness loss to reduce jittering.

Finally, as the ground plane location is not assumed to be known and HUND produces estimates in camera space $\kk$, we estimate the global transform  $\TT_g \in \RR^{3\times4}$ for the physical scene, with gravity along the y axis, as well as the ground plane at $y=0$. This is achieved by minimizing
\begin{equation}\label{eq:world_transform}
    L_g(\TT_g) = \sum_{i}^{N} \|\min\bigl(\delta, \LL_y\bigl(\TT_g[\MM(\bbeta,\btheta_i), 1]\bigr)\bigr)\|^2,
\end{equation}
where $\LL_y$ is an operator that extracts the $k=20$ smallest signed distances from the mesh vertices $\MM(\bbeta, \btheta_i)$ after the global transformation. This assumes the body is in ground plane contact for most of the sequence. To allow for frames where the subject is not in contact with the ground, we clip the maximum shortest distance to the ground to $\delta=20$ cm.

\subsection{Differentiable Physics Simulation Model}\label{sec:differentiable_physics}
\begin{figure}[bht]
  \centering
  \includegraphics[width=1.0\linewidth]{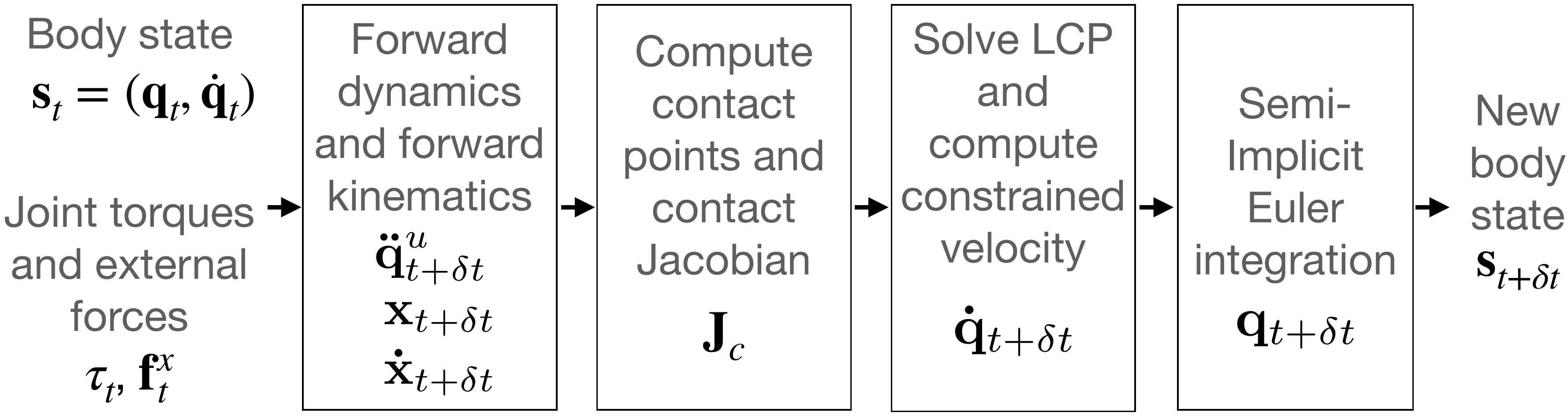}
  \vspace{-4mm}
  \caption{Overview of the simulation step of the physics model that updates the current state $S_t$ to a new state
    after time step $\delta_t$. For each computational block we include the output quantities used
    in the subsequent block.}
  \label{fig:tdsoverview}
  \vspace{-4mm}
\end{figure}

We implement our models in the framework of the ``Tiny Differentiable Simulator'' (TDS)~\citep{heiden2021neuralsim}. This formulates rigid-body dynamics for articulated bodies in terms of reduced coordinates. Elements in the vector $\qq$ represent the position of each joint, and elements in the vector $\qqd$ represent joint space velocities, based on revolute and spherical joints. Given the state of the body $\vec{s}_t = (\qq_t, \qqd_t)$ at time $t$, as well as the vector of joint
torques $\btau_t$, and external forces $\ff_t$, the computation shown in \Fig{fig:tdsoverview}
produces a new body state $\vec{s}_{t+\delta t}$ corresponding to the rigid multi-body dynamics with contacts.
To that end, we first run forward kinematics to compute world space positions and
velocities, as well as forward dynamics to compute unconstrained acceleration obtained without
taking contacts into account. The forward dynamics computes the acceleration by solving the equation
of motion for the kinematic tree given by
\begin{equation}
\btau_{t} = \HH(\qq_{t})\qqdd_{t} + \CC(\qq_{t}, \qqd_{t}, \ff_{t}^{x})
\end{equation}
where $\HH(\qq)$ is the joint-space inertia matrix, $\CC$ the a joint space bias force and $\ff^{x}$ is the
vector of external forces. The forward dynamics is computed by
propagation-based Articulated-Body Algorithm (ABA)~\citep{featherstonerbda} that traverses the kinematic chain of the body three times in order to compute quantities necessary to finally obtain the acceleration of each rigid component of the body\footnote{See tab.~7.1 in~\citet{featherstonerbda} for the Articulated-Body Algorithm.}. The joint-space inertia matrix is computed using the Composite Rigid Body Algorithm (CRBA)~\citep{featherstonerbda}.

Unconstrained accelerations $\qqdd_{t+\delta t}^u$ are then used to compute unconstrained velocities, which in conjunction with the output of the forward kinematics $\xx_{t+\delta t}$ are used to update the
contact points between the articulated body and the environment. Contact points with positive (separating) distance are classified as inactive, while contact points with zero or negative distance are active. Active contacts generate a repulsive impulse that needs
to be taken into account when computing the new body state. To that end, the forward dynamics
computation is phrased as a linear complementarity problem (LCP) at the velocity level \citep{stepien2013phd,stewardtrinkle2000}
\begin{align}
\label{eq:lcp}
  \mat{J}_c\HH^{-1}\mat{J}_c^\top\pp + \mat{J}_c\xxd & = \vv\\
   \vv = \left[\vv_u, \vv_b\right] & \nonumber \\
  \mbox{s.t.}\qquad \vv_u^{\top}\pp_u=0 \qquad \vv_u \geq 0 & \qquad \pp_u \geq 0 \qquad \vv_b = 0 \nonumber
\end{align}

where $\mat{J}_c$ is a contact Jacobian for the positions of contact points computed in the previous step, $\pp$ is the vector of reaction impulses, and $\vv$ is the vector of relative velocities. The indices $u$ and $b$ indicate the unilateral and bilateral portion of constraints, respectively. The LCP problem in \Eq{eq:lcp} is then iteratively solved with a projected Gauss-Seidel method following the formulation in \citep{stepien2013phd}, by relying on a per-contact LCP \citep{per_contact_iteration2018}. The final step of the computation is to obtain joint positions $\qq_{t+\delta t}$ from joint velocities using semi-implicit Euler integration.

\subsection{Physical Human Body Modeling}\label{sec:body_model}
In the physical simulation, we model the human body as rigid geometric primitives connected by joints. The model is comprised of 16 joints with a total of 48 degrees of freedom, joining together 26 capsules that represent the various body parts (\cf \Fig{fig:overview}). The shape and mass of the model is automatically adapted for various body shapes by relying on a statistical body model~\citep{xu2020ghum}. Given the 3d mesh $\MM(\bbeta, \vect{\emptyset})$ corresponding to a shape estimate $\bbeta$ in rest pose, we infer the dimensions of the geometric primitives following the approach of \citep{alborno18cgf}. The process is entirely automatic and yields individualized physical models for each subject. As a physical model requires mass, we first estimate the total mass of the body based on a human shape dataset~\citep{pishchulin17pr} then distribute the weight according to an anatomical distribution~\citep{plagenhoef1983anatomical}. Finally, the inertia of each primitive is computed based on its mass and dimensions.

DiffPhy reconstructs a motion in simulation by actuating torque motors in the joints of the body. Following prior work~\citep{alborno13vcg} we optimize over control targets to proportional-derivative (PD) controllers rather than over the torques directly. We define the body's angular joint positions as $\qq_t$, and joint velocities as $\qqd_t$, the associated 3d Cartesian coordinates of the joints as $\xx_t$ for the time step $t$. Given a set of joint targets $\hat\qq_{1:T} = \{\hat\qq_1, \hat\qq_2, \ldots, \hat\qq_t\}$ the PD controllers infer the joint torques as 
\begin{equation}\label{eq:pd_controller}
    \btau_{t} = k_p(\hat{\qq}_t - \qq_t) - k_d\qqd_t,
\end{equation}
where $k_p$ and $k_d$ are gain parameters of PD controllers. We may then specify a motion of length $T$ as the initial state $\vect{s}_0 = (\qq_{0}, \qqd_{0})$, the world geometry $\mat{G}$ defining the position and orientation of the
ground plane, and a target trajectory for the joints $\hat\qq_{1:T}$. Given the loss presented in \eqref{eq:physics_loss} we reconstruct the motion by minimizing $L = L(\vect{s}_0, \mat{G}, \hat\qq_{1:T})$ with respect to $\hat{\qq}_{1:T}$.

\subsection{Gradient-Based Optimization}\label{sec:optimization}
\begin{table}[t]
\begin{center}
\scalebox{0.73}{
\begin{tabular}{l|c|c|c|c|c}
\textbf{Method} & \textbf{\# eval} & \textbf{MPJPE-G} &\textbf{MPJPE} & \textbf{MPJPE-PA} & \textbf{MPJPE-2d} \\
\hline

CMA-ES & 80k & 206.7 & 125.7 & 77.4 & 16.9 \\
BFGS & 122 & 160.1 & 100.1 & 68.9 & 15.5 \\
Basin-BFGS & 509 & 144.9 & 84.6 & 61.1 & 12.6 \\
\end{tabular}
}
\end{center}
\vspace{-5mm}
\caption{Comparison of optimization strategies on our Human3.6M validation set. BFGS and Basin-BFGS both use gradients, while CMA-ES is a gradient-free approach. Note that Basin-Hopping together with BFGS (Basin-BFGS) improves the performance of BFGS by combining it with stochastic global optimization. Using only a purely sampling-based approach (CMA-ES) requires magnitudes more function evaluations while still not finding better optima for our loss. BFGS was given a sufficiently large evaluation budget to converge.}
\vspace{-2mm}
\label{tab:strategies}
\end{table}

Given our loss function $L = L(\vect{S}_0, \mat{G}, \hat\qq_{1:T})$ we can use any gradient-based optimization method to
minimize the loss with respect to $\hat\qq_{1:T}$. Since the loss function is non-convex, convergence to suboptimal local minima is possible. Therefore, a global optimization combined with a local gradient-based search is expected to outperform a purely local method. One such method is the global stochastic optimization \emph{Basin-Hopping}~\citep{wales1997basin}. It uses a two-stage approach, which alternates between performing gradient-based local search and stochastic global search. Based on an initial candidate, it first performs a local search. It then randomly perturbs the local minimum, performs a local search again on the new candidate, and then either accepts or rejects the new solution based on the Metropolis criterion~\citep{metropolis1953equation}. In our model, we use BFGS~\citep{Flet87} for local optimization.

\subsection{Optimization Objectives}\label{sec:objectives}
Reconstructing a motion sequence amounts to finding the control trajectory $\hat\qq_{1:T}$ that minimizes the reconstruction loss $L$ under the constraints of the simulation dynamics. In this work, we formulate $L$ as a weighted combination of loss functions
\begin{equation}\label{eq:physics_loss}
    L=w_r L_r + w_j L_j + w_i L_i + w_l L_l ,
\end{equation}
with the weights $w_r = 10.0$, $w_j = 0.1$, $w_i = 0.01$, and $w_l = 0.01$. The root position loss $L_r$ measures errors between the 3d position of the simulated pelvis root joint $\xx_{t}^\text{root}$ and the kinematically estimated position $\bar\xx_{t}^\text{root}$
\begin{equation}
L_{r}(\hat\qq_{1:T}) = \frac{1}{T}\sum_t^T\|\bar\xx_{t}^\text{root} - \xx_{t}^\text{root}\|^2 
\label{eq:rootloss}
\end{equation}
at time $t$ where $T$ is the total length of the sequence. $L_j$ computes the rotational distance between the kinematic pose estimate and the simulated body's pose
\begin{equation}
\label{eq:rotloss}
L_{j}(\hat\qq_{1:T}) = \frac{1}{TK}\sum_t^T\sum_{k}^K\arccos(|\qq^k_{t} \cdot \bar\qq^k_{t}|),
\end{equation}
where $\bar\qq_{t}^k$ and $\qq_t^k$ are rotations expressed as quaternions for joint $k$ at time $t$ for kinematics and the simulated character respectively. Note the difference between $\hat\qq$ and $\qq$, where the former are the PD control targets and the latter are the joint angles of the simulated model (\cf \Eq{eq:pd_controller}). $L_i$ computes the 2d projection loss 
\begin{equation}
L_{i}(\hat\qq_{1:T}) = \frac{1}{TK}\sum_t^T\sum_{k}^K \cc^k_t\|\bar\xx^k_{t} - \Pi(\xx^k_{t}, \ii)\|^2,
\label{eq:2dloss}
\end{equation}
where $\Pi(\xx^k_{t}, \ii)$ is the perspective operator projecting the simulated model's joint $\xx^k_t$ onto the image with camera intrinsics $\ii$ weighted by the keypoint detection confidence score $c^k_t$. Finally, $L_l$ is a regularizer that penalizes joints outside of human anatomical limits as present in the statistical body model~\citep{xu2020ghum}
\begin{equation}
\begin{split}
\label{eq:limitloss}
L_{l}(\hat\qq_{1:T}) = \frac{1}{TK}\sum_t^T\sum_{k}^K\| & \max(z^k_\text{lower} - \qq^k_t, 0) \\
& + \max(\qq^k_t - z^k_\text{upper}, 0) \|^2,
\end{split}
\end{equation}
where $z^k_{upper}$ and $z^k_{lower}$ are upper and lower bounds for joint $k$ respectively. 

Note that in the above definitions, the positions of body joints angles $\qq_{t}^k$ and 3d joint positions $\xx_{t}^k$ are dependent on the control trajectory up until time $t$, as part of the physics formulation introduced in \Sec{sec:differentiable_physics}. 

\subsection{Optimized Initialization}
\label{sec:optimized_initialization}
We initialize the pose $\qq_0$ in the first time step of the simulation to the kinematically estimated pose $\bar\qq_0$ and estimate the velocity $\qqd_0$ using finite differences between the first two kinematic poses \{$\bar\qq_0$, $\bar\qq_1$\}. However, if the initial kinematic pose estimate is poor, this might lead to a low quality starting pose from which the simulation cannot recover. Similarly, jitters in the kinematic poses may cause a significant error in the estimated initial velocity. We address these issues by including the initial pose and velocity as variables to optimize. We experimentally validate how such a relatively straightforward approach significantly impacts the results.

\section{Experiments}\label{sec:experiments}
\begin{figure*}[tb]
  \centering
  \includegraphics[width=0.9\linewidth]{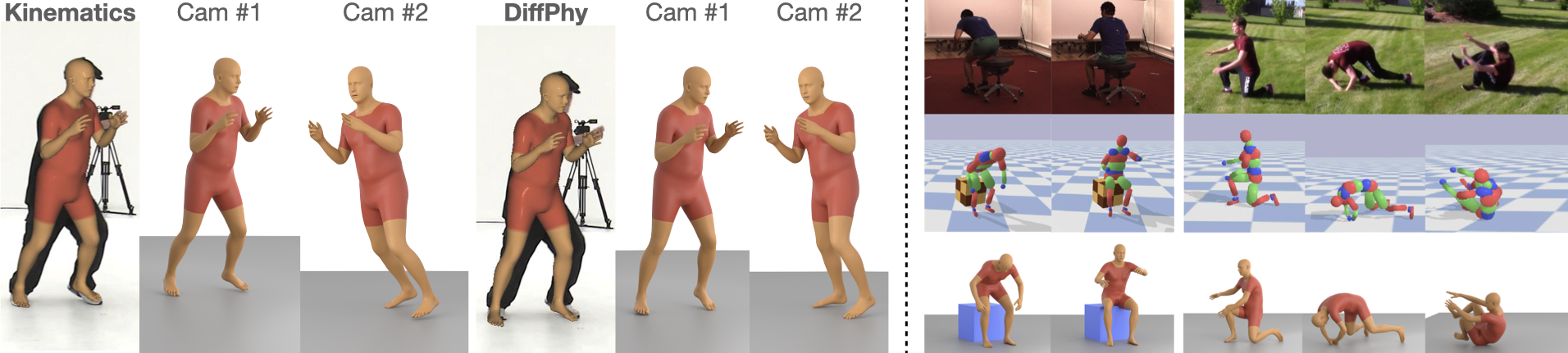}
  \vspace{-2mm}
  \caption{Qualitative examples on the AIST dataset (left) and of complex contacts (right). The AIST example shows that both kinematics and DiffPhy projects well into the image. However, when rendered from another viewpoint (cam \#2) it becomes clear that kinematics exhibits unrealistic leaning while the physical constraints corrects the pose to keep the body in balance. See \href{https://tiny.cc/diffphy}{tiny.cc/diffphy} for more.}
  \vspace{-0mm}
  \label{fig:aist_qualatative}
\end{figure*}

\begin{table*}[htbp]
\begin{center}
\scalebox{0.90}{
\begin{tabular}{c|l|c|c|c|c|c|c}
\textbf{Dataset} & \textbf{Model} & \textbf{MPJPE-G} & \textbf{MPJPE} & \textbf{MPJPE-PA} & \textbf{MPJPE-2d} & \textbf{TV} & \textbf{Foot skate (\%)} \\
\hline
\multirow{7}{*}{Human3.6M} & VIBE~\cite{kocabas20cvpr} & 207.7 & 68.6 & 43.6 & 16.4 & 0.32 & 27.4 \\
& PhysCap~\cite{PhysCapTOG2020} & - & 97.4 & 65.1 & - & - & -  \\
& SimPoE~\cite{yuan2021simpoe} & - & \textbf{56.7} & \textbf{41.6} & - & - & -  \\
& Shimada \etal~\cite{Shimada2021NeuralM3} & - & 76.5 & 58.2 & - & - & -  \\
& Xie \etal~\cite{Xie_2021_ICCV} & - & 68.1 & - & - & - & -  \\
& Kinematics & 145.3 &  83.0 & 55.4 & 13.4 &  0.34 & 47.5 \\
& DiffPhy & \textbf{139.1} & 81.7 & 55.6 & \textbf{13.1} & \textbf{0.20} & \textbf{7.4}  \\

\hline
\multirow{2}{*}{AIST} 
& Kinematics & 155.7 & 107.4 & 66.9 & \textbf{10.4} & 0.52 & 50.9 \\
& DiffPhy  &  \textbf{150.2} & \textbf{105.5} & \textbf{66.0} & 12.1 & \textbf{0.44} & \textbf{19.6}  \\
\end{tabular}
}
\end{center}
\vspace{-5mm}
\caption{Quantitative evaluation on the Human3.6M and AIST datasets. Our full dynamic model improves over the kinematic estimates used as initialization with respect to standard joint position error metrics as well as reducing motion jitter and unnatural foot skating.}
\label{tab:quantitative}
\vspace{-4mm}
\end{table*}
\begin{figure*}[tb]
  \centering
  \includegraphics[width=0.86\linewidth]{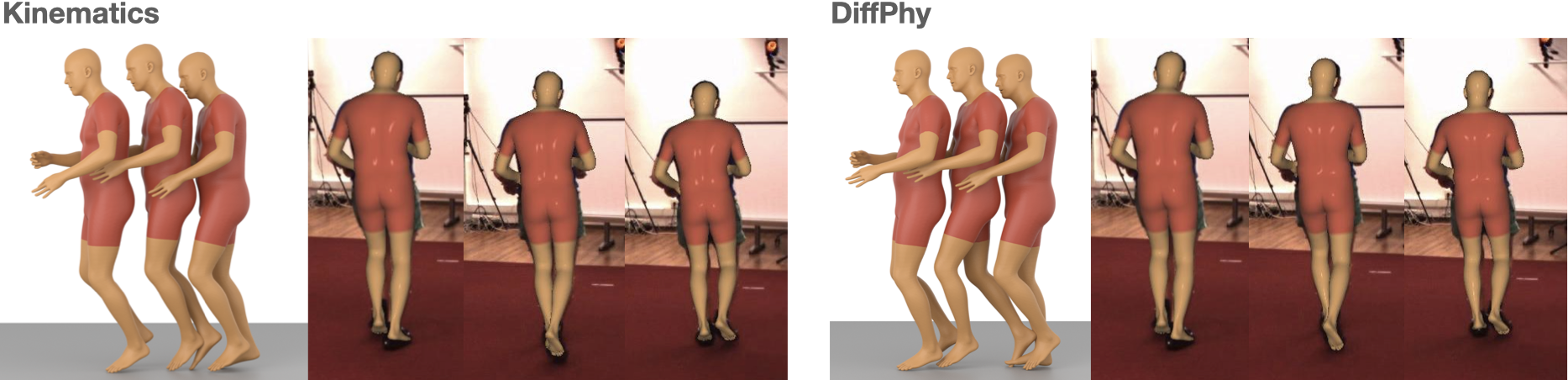}
  \vspace{-2mm}
  \caption{Qualitative examples on Human3.6M. DiffPhy infers plausible leg motion while kinematics skates unrealistically forward.}
  \label{fig:h36m_qunatalitative}
  \vspace{-4mm}
\end{figure*}

\myparagraph{Datasets.}\label{par:datasets}
We quantitatively evaluate DiffPhy on the Human3.6M~\cite{h36mpami}, and a subset of the AIST~\cite{aist-dance-db} pose datasets. The former contains a diverse set of motions from a motion capture laboratory, whereas the latter contains dance videos with triangulated 3d joints as pseudo-ground-truth. As only DiffPhy and SimPoe~\cite{yuan2021simpoe} supports full-body contacts (\cf \Tab{tab:competitors}), PhysCap~\cite{PhysCapTOG2020} proposed evaluating on a subset of the Human3.6M. This protocol eliminated all sequences requiring more than foot-floor contacts. Hence to allow for comparison, we use this subset in \Tab{tab:quantitative}, but note that our method is more general and supports contacts for all body parts. For ablations, we use 100 frames from 20 sequences from a validation subset of Human3.6M. Finally, we quantitatively evaluate our method on real-world internet videos released under creative commons licenses. For additional details, refer to our supplementary material.

\myparagraph{Metrics.} We report the standard pose metrics such as mean per-joint position error in millimeter (MPJPE-G), mean Procrustes aligned joint error (MPJPE-PA), per-frame translation aligned error (MPJPE), and 2d mean per-joint error in pixels (MPJPE-2d). Note that many papers do not report global position errors since they consider only root-relative poses. We, however, are interested in measuring the pose error, including translational errors, since unnatural translation is a common (non-physical) reconstruction artifact. In addition, we also measure foot skating and the total variation in the joint acceleration per frame (TV). We measure foot skating as percentage of frames where a foot moves more than $2$cm while in contact with the ground in two adjacent frames. Unlike \cite{RempeContactDynamics2020}, we do not assume foot contact annotations but instead heuristically detect foot contacts based on the distance between the foot mesh and the ground-plane. The total variation in acceleration is computed as $\frac{1}{T}\sum_{t\in T}\sum_{k \in K} |\ddot{x}^k_{t+1} -\ddot{x}^k_{t}|$, for the 3d acceleration $\ddot{x}^k_t$ of joint $k$ at time $t$ estimated using finite differences. Thus, high TV indicates motion jitter, and high foot skate implies motion that slides along the ground.

\myparagraph{Implementation Details.}\label{par:implementation}
We use the Tiny Differentiable Simulator~\cite{heiden2021neuralsim} running at $1,000$ Hz with the gradients computed using the auto differentiation framework CppAD~\cite{cppad}. In addition, we use a Python implementation of Basin-Hopping and BFGS~\cite{2020SciPy-NMeth}. Since the length of the optimized trajectory may be great, we follow \cite{alborno13vcg} and perform optimization in overlapping windows of length $N=960$. The simulation steps take $\approx 5$s. For the large datasets in \Tab{tab:quantitative}, we compute the windows in parallel and stitch them together in order to speed up computation. We initialize the control targets $\hat{q}_{1:T}$ to 3d poses estimated by our kinematics. See our supplementary material for details.

\subsection{Results}
We compare DiffPhy against both state-of-the-art kinematic video models (VIBE~\cite{kocabas20cvpr}) and against physics-based methods. The results are summarized in \Tab{tab:quantitative}. Since VIBE predicts root-relative poses, we estimate the global translation (required to compute MPJPE-G) by minimizing 2d projection errors using a method similar to the one in \cite{PhysCapTOG2020}. For VIBE, we use the publicly available implementation. For the other methods, we give numbers presented by the authors. 
On both Human3.6M and AIST, our model improves with respect to the physical metrics (TV and foot skate) compared to the kinematic initialization. On Human3.6M, foot skating is only $7.4$\% compared to $47.5$\% for the kinematic initialization and $27.4$\% for VIBE. On AIST, foot skating is reduced from $50.9$\% to $19.6$\%. We believe that increased skating on AIST is due to actual skating motions performed as part of the hip-hop dances. On total variation, our model similarly improves over kinematics with $0.20$ and $0.44$ on Human3.6M and AIST, respectively.
Furthermore, we note that our full model improves the global joint position error (MPJPE-G), a metric that measures pose and translation errors. On Human3.6M, DiffPhy has an error of $139.1$ compared to $145.3$ and $207.7$ mm/joint for kinematics and VIBE, respectively. If we look at the error for foot joints only, we see an even larger improvement by including physics compared to kinematics alone ($166.8$ vs. $174.1$ mm/joint). This result aligns with prior work~\cite{RempeContactDynamics2020}, showing that physics improves foot position estimation. Furthermore, our method aligns well with image evidence when comparing 2d error, i.e., $13.1$ px/joint vs. VIBE's $16.4$ px/joint on Human3.6M. In terms of joint error including translation error (MPJPE), SimPoE~\cite{yuan2021simpoe}, Xie \etal~\cite{Xie_2021_ICCV}, Shimada \etal~\cite{Shimada2021NeuralM3} outperform DiffPhy ($56.7$ vs. $68.1$ vs. $76.5$ vs. $81.7$ mm/joint respectively), though in the case of SimPoE and Xie \etal this might stem from initializing from the already strong VIBE predictor ($68.6$ mm/joint). Furthermore, SimPoE is a neural network requiring extensive training using the 3d ground-truth from Human3.6M, whereas DiffPhy is a general method that requires no additional training (\cf \Tab{tab:competitors}). Xie \etal, PhysCap~\cite{PhysCapTOG2020}, Shimada \etal~\cite{Shimada2021NeuralM3} on the other hand, focus only on feet-ground contacts while DiffPhy supports complex contacts. Unfortunately, this advantage cannot be demonstrated on subsets that exclude sequences with complex contacts. 

\FFig{fig:h36m_qunatalitative} presents qualitative results where kinematics fails to estimate the positions of legs due to depth ambiguities. The reconstructed poses align well when projected into the image but are unrealistic since the model skates rather than walks forward. Since DiffPhy reconstructs the motion with physics in the loop, it must propel the model forward through bipedal locomotion, thus inferring feasible leg poses. Similarly, in \Fig{fig:aist_qualatative} kinematics estimates a pose that projects well into the image. However, when viewed from a side, it becomes clear that kinematics estimates a pose that leans unnaturally. Since DiffPhy is constrained by gravity, it must find a pose that is both physically plausible \emph{and} aligns with 2d evidence. \FFig{fig:aist_qualatative} also includes examples of object interactions and rolling motions requiring complex contacts. We manually modeled the chair as a box since DiffPhy does not estimate scene geometry. For the rolling motion, the kinematics were too noisy for DiffPhy to converge; hence, we manually corrected the worst kinematic frames before running DiffPhy. Finally, \Fig{fig:pim_qunatitative} shows two reconstruction examples for sequences in-the-wild. These videos exhibit poses and activities missing from standard laboratory-captured datasets.

\myparagraph{Ablation studies.} In \Tab{tab:loss_ablation} we validate our choice of loss components in \Eq{eq:physics_loss}. We note that the 2d projection loss, as expected, plays an important role in aligning the reconstruction with the image evidence ($17.1$ vs. $12.6$ px/joint). Furthermore, since 2d keypoints do not suffer from depth ambiguities, they are generally more reliable than 3d keypoints and thus serve as a strong signal. Therefore removing 2d evidence significantly increases MPJPE-G from $144.9$ to $158.5$ mm/joint. Removing the root position loss \Eq{eq:rootloss} has the largest impact on global position error ($165.7$ mm/joint) since without it, we do not provide DiffPhy with any supervision with respect to world positioning. This allows for suboptimal reconstructions that align well with the projected image ($12.8$ px/joint) but do not transition correctly in world space. Without the joint angle loss \Eq{eq:rotloss}, DiffPhy is deprived of the per-frame 3d pose estimates, which, when predicted by neural networks such as HUND or VIBE that are trained on large pose datasets, provide useful guidance as long as their predictions do not contradict any physical constraints. Removing the joint angle limit regularizer (\cf \Eq{eq:limitloss}) demonstrates the usefulness of constraining the reconstructed motion to the space of anatomically valid poses even for everyday motions like those in Human3.6M. Finally, we validate the usefulness of optimizing the initial starting pose and velocity (see \Sec{sec:optimized_initialization}). Without it, the kinematic estimates for the initial frames must be accurate. If not, the simulation may start from an initial state from which DiffPhy may fail to recover, as seen by the largest MPJPE-PA in the ablation of $65.1$ mm/joint.

Next, results in \Tab{tab:strategies} show that gradient-based methods are vastly more efficient for our physics loss compared to the commonly used gradient-free approach CMA-ES~\cite{Hansen2006}. BFGS obtains a lower MPJPE-G error ($160.1$ vs $206.7$ mm/joint), and requires a fraction of the computations ($122$ vs. $80$k loss evaluations per windows). Next, We note that BFGS converges to suboptimal minima, but by combining BFGS with Basin-Hopping, we can reduce the errors further to $144.9$ mm/joint. As Basin-Hopping can explore infinitely many basins, we set the limit to 5 basin steps, each with 50 BFGS iterations as a trade-off between accuracy and speed.

In \Tab{tab:window_size} we study the effect of the optimization window size. We find that a window of $960$ simulation steps (containing $0.96$s of video) is optimal for our setup. A larger window size increases the errors, most likely due to a larger search space combined with a larger gradient variance, as noted in \cite{metz2021gradients}. On the other hand, smaller windows provide scarcer visual evidence and are sensitive to a few occluded frames, or to noisy estimates. Interestingly, a smaller window size performed better for experiments on ground-truth data (see supplementary material). This indicates that smaller apertures are better for noise-free inputs.

Several methods (\cf \Tab{tab:competitors}) introduce ``residual forces'' acting on the root link of the physical body. This non-physical force allows the method to translate and rotate the body to align with visual evidence at the expense of physical realism. \Tab{tab:root_force} confirms that this indeed can be used to lower DiffPhy's joint errors (MPJPE-G from $144.9$ to $140.2$ mm/joint and 2d error from $12.6$ to $11.6$ px/joint when applying $50$N for each of the six degrees of freedom). Interestingly, applying a too great residual force ($100$N) increased error, perhaps since it allows the model to circumvent some of the constraints of physical simulation. In this work, we avoid using residual forces, in order to keep all forces realistic, and avoid non-physical artifacts.

\begin{table}[bt]
\begin{center}
\scalebox{0.85}{
\begin{tabular}{l|c|c|c|c}
\textbf{RF} & \textbf{MPJPE-G} &\textbf{MPJPE} & \textbf{MPJPE-PA} & \textbf{MPJPE-2d} \\
\hline
0 & 144.9 & 84.6 & 61.1 & 12.6 \\
5 & 141.4 & 82.2 & 60.7 & 11.8 \\
10 & 140.1 & 79.9 & 60.2 & 11.7 \\
25 & 146.3 & 81.9 & 60.0 & 12.7 \\
50 & 140.2 & 79.4 & 60.3 & 11.6 \\
100 & 154.0 & 87.7 & 61.5 & 14.4 \\
\end{tabular}
}
\end{center}
\vspace{-4mm}
\caption{Results on experiments on the effects of residual force. We note using a residual force decreases the error metrics, but we refrain from using it to avoid unexplained non-physical forces.}
\vspace{-0mm}
\label{tab:root_force}
\end{table}

\begin{table}[bt]
\begin{center}
\scalebox{0.85}{
\begin{tabular}{l|c|c|c|c}
\textbf{Window} & \textbf{MPJPE-G} &\textbf{MPJPE} & \textbf{MPJPE-PA} & \textbf{MPJPE-2d} \\
\hline
240 & 390.1 & 224.1 & 96.6 & 40.3 \\
480 & 165.6 & 97.2 & 63.8 & 13.2 \\
720 & 148.9 & 87.2 & 61.8 & 12.6 \\
960 & 144.9 & 84.6 & 61.1 & 12.6 \\
1440 & 155.6 & 92.5 & 65.7 & 15.9 \\
\end{tabular}
}
\end{center}
\vspace{-6mm}
\caption{Results on the effects of  optimization window size. A balance needs to be found between a larger window  size which allow for more visual evidence to be taken into account while a smaller reduces the dimensionality of the search space.}
\vspace{-4mm}
\label{tab:window_size}
\end{table}
\begin{table}[bt]
\begin{center}
\scalebox{0.85}{
\begin{tabular}{l|c|c|c|c}
\textbf{Variant} & \textbf{MPJPE-G} &\textbf{MPJPE} & \textbf{MPJPE-PA} & \textbf{MPJPE-2d} \\
\hline
Full model & 144.9 & 84.6 & 61.1 & 12.6 \\
No root & 165.7 & 84.8 & 60.7 & 12.8 \\
No 2d & 158.5 & 98.3 & 65.7 & 17.1 \\
No pose & 156.8 & 91.8 & 64.4 & 13.0 \\
No 3d loss & 216.6 & 122.4 & 76.3 & 12.6 \\
No limits & 146.8 & 86.5 & 62.2 & 13.0 \\ 
No opt. init. & 151.5 & 92.1 & 65.1 & 14.1 \\
\end{tabular}
}
\end{center}
\vspace{-4mm}
\caption{Ablation of the model components introduced in \Sec{sec:methodology}. \emph{No root} means without root position loss \Eq{eq:rootloss}, \emph{No 2d} without 2d keypoint loss \Eq{eq:2dloss}, \emph{No pose} without joint angle loss \Eq{eq:rotloss}, \emph{No 3d loss} without both root link position loss and joint angles losses, \emph{No limit} without anatomical joint limits \Eq{eq:limitloss}, and \emph{No init. opt.} is without optimizing the initial state, \cf \Sec{sec:optimized_initialization}.}
\vspace{-2mm}
\label{tab:loss_ablation}
\end{table}

\section{Discussion}
In order to improve the realism of 3d human sensing, we have introduced \emph{DiffPhy} -- the first differentiable physics-based model for full-body articulated human motion estimation, that supports complex contacts, does not assume a known ground plane, and avoids reliance on non-physical forces. This has the benefit of a human model with realistic physics interactions, that are constrained end-to-end by visual losses. Furthermore, such a model can provide a valuable non-learning-based component, which is always valid, complementing the statistical kinematic prediction and optimization techniques prevalent in the current state of the art. Visual 3d human motion reconstruction experiments on multiple datasets demonstrate that our methodology is competitive with other state of the art physics-based approaches. 

\myparagraph{Limitations and Future Work.} An inherent limitation to physics-based approaches is the need to model objects in the scene. We hope to address this challenge in future work by integrating with 3d scene reconstruction techniques \cite{behave-cvpr22}. Ideally, we would be able to jointly optimize the control of the body and the world to match visual evidence. Another limitation is our current assumption of constant camera extrinsics. This limits our technique to videos captured using a static camera but can be easily relaxed. Finally, our reconstructions are limited to a single subject. Reconstructing multiple people interacting is interesting since these scenes are complex, and learning statistical models of interaction between humans is challenging~\cite{fieraru2021remips}. A physics-based approach could help infer constraints and affordances.

\myparagraph{Ethical Considerations.} Our construction of physics-based models is motivated by the breadth of transformative 3d applications that would become possible, including fitness, personal well-being or special effects, or human-computer interaction, among others. In contrast, applications like visual surveillance and person identification would not be effectively supported, given that the model's output does not provide sufficient detail for these purposes. The same is true for the creation of potentially adversely-impacting deepfakes, as an appearance model or a joint audio-visual model are not included for photorealistic visual and voice synthesis. While our method is fundamentally applicable to a variety of human body types, we have not evaluated this aspect extensively and consider such a study an important objective for future work.

\clearpage
{\small
\bibliographystyle{ieee_fullname}
\bibliography{biblio}
}

\clearpage
\appendix
\section*{Appendix}
This supplement presents additional results (\Sec{sec:sm_results}), a description of the datasets used (\Sec{sec:sm_dataset}) together with a description of the usage of data with human subjects (\Sec{sec:human_subjects}), and additional details of the simulation setup (\Sec{sec:sm_diff_physics}). Please refer to our video for qualitative results at \href{https://tiny.cc/diffphy}{tiny.cc/diffphy}. 

\section{Additional Results}\label{sec:sm_results}

\TTab{tab:window_size_mocap} presents an ablation on window size performed using mocap data as initialization and reference trajectory rather than using the kinematic initialization. In this case, we note that a smaller window size of 480 outperforms the larger window size of 960 used in the main paper. We hypothesize that when the reference signal lacks noise, a smaller window is easier to optimize since the dimension of the problem is reduced. However, with noisy observations, a larger window is required for the method to be robust to missing or poor  kinematic reconstructions.

\begin{table}[hbt]
\begin{center}
\scalebox{1}{
\begin{tabular}{l|c|c|c}
\textbf{Window} & \textbf{MPJPE-G} &\textbf{MPJPE} & \textbf{MPJPE-PA} \\
\hline
240 & 112.8 & 75.9 & 40.1 \\
480 & 39.4 & 33.4 & 21.9  \\
720 & 46.1 & 42.1 & 29.4  \\
960 & 77.8 & 68.4 & 44.9  \\
\end{tabular}
}
\end{center}
\caption{Ablation study of the optimization window size. Experiments were carried out on motion capture rather than the kinematic initialization as input. The experiment was performed on the same Human3.6M sequences as in the ablation in the main paper. Note that when using mocap rather than noisy observations, a smaller window size is better (480 vs. 960 in main paper).}
\label{tab:window_size_mocap}
\end{table}

\section{Datasets}\label{sec:sm_dataset}

\begin{table}[ht]
    \centering
    \begin{tabular}{c|c|c|c}
        \textbf{Sequence} & \textbf{Subject} &
        \textbf{Camera Id} & \textbf{Frames}  \\
        \hline
        Phoning & S11 & 55011271 & 400-599 \\
        Posing\_1 & S11 & 58860488 & 400-599 \\
        Purchases & S11 & 60457274 & 400-599 \\
        SittingDown\_1 & S11 & 54138969 & 400-599 \\
        Smoking\_1 & S11 & 54138969 & 400-599 \\
        TakingPhoto\_1 & S11 & 54138969 & 400-599 \\
        Waiting\_1 & S11 & 58860488 & 400-599 \\
        WalkDog & S11 & 58860488 & 400-599 \\
        WalkTogether & S11 & 55011271 & 400-599 \\
        Walking\_1 & S11 & 55011271 & 400-599 \\
        Greeting\_1 & S9 & 54138969 & 400-599 \\
        Phoning\_1 & S9 & 54138969 & 400-599 \\
        Purchases & S9 & 60457274 & 400-599 \\
        SittingDown & S9 & 55011271 & 400-599 \\
        Smoking & S9 & 60457274 & 400-599 \\
        TakingPhoto & S9 & 60457274 & 400-599 \\
        Waiting & S9 & 60457274 & 400-599 \\
        WalkDog\_1 & S9 & 54138969 & 400-599 \\
        WalkTogether\_1 & S9 & 55011271 & 400-599 \\
        Walking & S9 & 58860488 & 400-599 \\
    \end{tabular}
    \caption{Human3.6M~\cite{h36mpami} sequences used for ablation studies. Note that we downsampled the sequences from 50 FPS to 25 FPS.}
    \vspace{-2mm}
    \label{tab:h36m_sequences}
\end{table}
We evaluate our method on the two established datasets Human3.6M~\citep{h36mpami} and AIST~\citep{aist-dance-db}. In addition, we evaluate our method on ``real-world'' internet videos.

\myparagraph{Human3.6M.} When comparing to the state-of-the-art methods, we evaluate on the Human3.6M Protocol P2 sequences while excluding the same sequences as by Xie et al.~\cite{Xie_2021_ICCV}. That leaves the sequences: \textit{Directions, Discussions, Greeting, Posing, Purchases, Taking Photos, Waiting, Walking, Walking Dog and Walking Together}. We evaluate the motions using only camera \emph{60457274}. Similar to \cite{Xie_2021_ICCV}, we down sample the Human3.6M data from 50 FPS to 25 FPS.

The ablation studies were performed on a smaller subset of four-second clips (frames 400-599) from a random camera, see \Tab{tab:h36m_sequences}.

\myparagraph{AIST.} AIST provides dynamic dance motions not present in Human3.6M. We evaluate our method using the pseudo-ground-truth provided by \cite{li2021aistplusplus}. We use the first four seconds (120 frames) using a randomly selected camera from the sequences in \Tab{tab:aist_sequences}.

\myparagraph{Internet Videos.} Finally, we perform qualitative evaluation of our method on internet videos made public under creative common licences. %

\subsection{Metrics}
\myparagraph{Total variation.} We compute the total variation of the 3d joint acceleration as a measurement of the jitter in motion. This is given as
\begin{equation}
    \frac{1}{T}\sum_{t\in T}\sum_{k \in K} |\ddot{x}^k_{t+1} -\ddot{x}^k_{t}| ,
\end{equation}
where $\ddot{x}^k_{t}$ is the 3d joint acceleration of joint $k$ at time $t$. We estimate the acceleration through finite differences.

\myparagraph{Foot skating.} We track unnatural foot skating artifacts by measuring the percentage of frames where either foot is ``skating'' along the ground. Our formulation doesn't rely on foot contact annotations but instead heuristically detect when foot contacts occur by measuring the distance between the foot mesh and the ground-plane. A contact is defined as $N=10$ foot mesh vertices being within $d$ mm of the ground-plane. For kinematics we use $d=5$ mm and for dynamics $d=1$ mm to account for the capsule approximation being smaller than the foot mesh. We define skating as a foot moving $\geq 2$ cm between two frames while being in contact with the ground.

\subsection{Usage of data with human subjects} \label{sec:human_subjects}
In this work, we employ two established pose benchmarks that are commonly used in the field of human pose estimation. Human3.6M~\cite{h36mpami} was recorded in a laboratory setting with the permission of the actors, and AIST~\cite{aist-dance-db} contains \emph{``a shared database containing original street dance videos with copyright-cleared dance music. This is the first large-scale shared database focusing on street dances to promote academic research regarding Dance Information Processing''}\footnote{\url{https://aistdancedb.ongaaccel.jp/}}. As for the ``in-the-wild`` videos, these were released under creative common licenses granting express permission to \emph{``copy and redistribute the material in any medium or format''} and \emph{``remix, transform, and build upon the material for any purpose, even commercially''}. Finally, we do \emph{not} intend to release these videos as part of a dataset. Instead we only use them to demonstrate our method on videos with poses and motion uncommon in laboratory captured datasets.

\begin{table}[htb]
     \centering
     \begin{tabular}{c|c}
         \textbf{Sequence} & \textbf{Frames}  \\
         \hline
gBR\_sBM\_c06\_d06\_mBR4\_ch06 & 1-120 \\
gBR\_sBM\_c07\_d06\_mBR4\_ch02 & 1-120 \\
gBR\_sBM\_c08\_d05\_mBR1\_ch01 & 1-120 \\
gBR\_sFM\_c03\_d04\_mBR0\_ch01 & 1-120 \\
gJB\_sBM\_c02\_d09\_mJB3\_ch10 & 1-120 \\
gKR\_sBM\_c09\_d30\_mKR5\_ch05 & 1-120 \\
gLH\_sBM\_c04\_d18\_mLH5\_ch07 & 1-120 \\
gLH\_sBM\_c07\_d18\_mLH4\_ch03 & 1-120 \\
gLH\_sBM\_c09\_d17\_mLH1\_ch02 & 1-120 \\
gLH\_sFM\_c03\_d18\_mLH0\_ch15 & 1-120 \\
gLO\_sBM\_c05\_d14\_mLO4\_ch07 & 1-120 \\
gLO\_sBM\_c07\_d15\_mLO4\_ch09 & 1-120 \\
gLO\_sFM\_c02\_d15\_mLO4\_ch21 & 1-120 \\
gMH\_sBM\_c01\_d24\_mMH3\_ch02 & 1-120 \\
gMH\_sBM\_c05\_d24\_mMH4\_ch07 & 1-120
     \end{tabular}
     \caption{AIST~\cite{aist-dance-db} sequences used for evaluation.}
 \label{tab:aist_sequences}
\end{table}

\section{Differentiable Physics for Human Motion}\label{sec:sm_diff_physics}
Tiny Differentiable Simulator (TDS)~\cite{heiden2021neuralsim} is a C++ simulator where the data type is templetized. In our experiments, we use the scalar from the automatic differentiation (AD) framework CppAD~\citep{cppad} to compute the simulation gradients. That is, we compute the gradients of the loss with respect to the input control variables at each time step:

\begin{equation}
\frac{\partial L}{\partial \hat\qq_{1:T}} = \frac{\partial L}{\partial \qq_{1:T}} \frac{\partial \qq_{1:T}}{\partial \btau_{1:T}} \frac{\partial \btau_{1:T}}{\partial \hat\qq_{1:T}},
\label{eq:deriv}
\end{equation}

where $L$ is objective function of the trajectory optimization, $\qq_{1:T}$ are the simulated body's joint positions, and $\hat\qq_{1:T}$ are the per-timestep control signal to the PD controllers in the body joints.

To speed up the optimization we implement our simulation as a fixed computational graph of the simulation rollout for a fixed number of steps and then repeatedly use it to compute the values of the gradients in \Eq{eq:deriv}. This greatly speeds up the optimization since the automatic  differentiation framework doesn't need to setup the computational graph for each backward pass. To that end, we make the following adaptations to TDS to make it support a fixed graph.

\myparagraph{Differentiation and contact points.} Since at the time of graph construction it is not
known in advance which contact points will be active for particular inputs we always include all 
contact points into the LCP formulation. This increases the graph size based on the number of contacts considered. The issue of large graph can be address by e.g. ``checkpointing'' the computation as described in \cite{qiao2021efficient}. 

\myparagraph{Dealing with exploding gradients.} As noted in ~\cite{metz2021gradients}, gradients from differentiable simulators may explode or vanishing when the window size is large. In this work, we experimentally found it possible to mitigate the issue by setting the LCP solver iterations to $K=1$ without noticeable degradation of reconstruction quality.

\myparagraph{Implementation Details}
In our experiments we run TDS with a step size of $1$ms. This is partly due to the simpler PD controller, which requires smaller simulation steps to allow for stable control. We set the ground-plane friction to $0.8$ and the controller gains to $k_p=200$ and $k_d=5$. %
Evaluating our loss function and computing the gradients for a window of 960 simulation steps takes approximately $\approx 5$ seconds on a standard desktop computer with only feet contacts enabled. Enabling more contacts or simulating multiple objects increases memory and computation time.

\end{document}